%% file: main.tex
\documentclass[10pt]{cai}
\usepackage[table]{xcolor}
\usepackage{makecell}
\usepackage{graphicx}
\usepackage{svg}
\usepackage[export]{adjustbox}
\usepackage{pgfplots,pgfplotstable}
\usepgfplotslibrary{polar}
\usepackage{subcaption}
\usepackage{fancybox}
\pgfplotsset{compat=1.18}
\begin{document}
% Editorial staff will replace the following values:
% 1. Conference Year
% 2. Issue number
% 3. Article DOI
\def\conferenceyear{2026}
%\volumeheader{37}{0}%{00.000}
\begin{center}

%\title{Data-Efficient Few-Shot Cross-Domain Glioma Magnetic Resonance Imaging Synthesis}
\title{Anatomically-conditioned Latent Diffusion Model for Data-Efficient Few-Shot Cross-Domain 3D Glioma MRI Synthesis}
\maketitle

\thispagestyle{empty}

% Add Authors and Affiliations in the camera ready
% for the double blind review, please leave this section as is 
\begin{tabular}{cc}
Salman Shaik\upstairs{\affilone,*}, Truong Thanh Hung Nguyen\upstairs{\affilone,*}, Hung Cao\upstairs{\affilone}
\\[0.25ex]
{\small \upstairs{\affilone}Analytics Everywhere Lab, University of New Brunswick, Canada} \\
\end{tabular}
  
% Replace with corresponding author email address
\emails{
  \upstairs{*}salmanbasha.shaik@unb.ca, hung.ntt@unb.ca

}
\vspace*{0.2in}
\end{center}

\begin{abstract}
\input{sec/0_abs}
\end{abstract}

% add your keywords
%\copyrightnotice

\input{sec/1_intro}

\input{sec/2_rw}
\input{sec/3_method}

\input{sec/4_exp}

\input{sec/5_res}
\input{sec/6_conc}

\section*{Acknowledgment}
This work was supported by NSERC Discovery Grant No RGPIN-2025-04478 and NSERC Discovery Supplement Award No DGECR-2025-00129.

% All references should be stored in the file "references.bib".
% That call to use that file is in "cai.cls". 
% Please do not modify anything below this line.
\printbibliography[heading=subbibintoc]
\newpage
\input{sec/7_apdx}
\end{document}

%% file: sec/0_abs.tex
Accurate classification of diffuse gliomas is often hindered by domain shifts across centers and a lack of large, annotated datasets. We propose the \textit{Anatomically-conditioned Latent Diffusion Model (ALDM)}, a novel framework for data-efficient, few-shot 3D volumetric MRI synthesis. ALDM utilizes a two-stage approach: a 3D variational autoencoder learns anatomical priors from a data-rich source domain, while a conditional latent diffusion model, guided by tumor masks via a ControlNet, generates structurally coherent volumes for a data-scarce target domain. Evaluated in an extreme few-shot setting with only 16 target images, ALDM outperformed GAN and hybrid baselines, achieving a superior Fréchet Inception Distance (FID) of 85.40 and a downstream classification AUC of 0.987. Qualitative results confirm that the model preserves sharp pathology boundaries and cross-modal consistency, with visual fidelity improving progressively during training. By capturing essential diagnostic features, ALDM provides a robust tool for clinical data augmentation in low-resource settings. Our implementation is available at \url{https://github.com/Analytics-Everywhere-Lab/anatomically-conditioned-LDM}.

%% file: sec/1_intro.tex
\section{Introduction}
Diffuse gliomas are heterogeneous malignant brain tumors with varied prognoses, so accurate classification of grade and molecular subtype (e.g., IDH mutation, 1p19q codeletion) is important for treatment planning \cite{calabrese2022university,bakas2021multi}. Deep learning (DL) models can classify gliomas from multimodal MRI with high accuracy on single-institution data, but performance often drops under multicenter domain shift and in limited-sample settings due to the lack of large, consistently annotated datasets \cite{roca2025iguane,kamnitsas2017unsupervised,fortin2018harmonization,bento2022deep}. Radiological heterogeneity further complicates generalization. WHO Grade IV glioblastoma, particularly IDH-wildtype, often shows strong contrast enhancement and elevated cerebral blood volume, whereas WHO Grade II-III and IDH-mutant gliomas more often exhibit weaker enhancement and reduced perfusion \cite{suh2019imaging}. These IDH-linked differences induce a clinically meaningful domain shift that single-domain synthesis methods do not adequately address.

Acquiring high-quality multimodal brain MRI is resource-intensive because protocols require multiple sequences and one hour of scan time. In routine practice, some sequences are missing due to limited time or image artifacts. Across institutions, differences in scanner vendor, field strength, coils, and protocol settings introduce site effects that shift image statistics and can reduce the external performance of DL models \cite{bento2022deep}. Generative Adversarial Network (GAN) and conditional Latent Diffusion Models (LDM) have shown strong synthesis quality and better diversity than GANs, and data-efficient fine-tuning can adapt diffusion generators using limited target data \cite{dorjsembe2024conditional,truong2024synthesizing}. However, most glioma synthesis studies focus on two-dimensional (2D) slices and are rarely evaluated in few-shot cross-domain settings \cite{na2025laplacian,mukherkjee2022brain}. High-quality target-domain volumetric synthesis could reduce the need for additional scans and help mitigate dataset shift across sites.

This motivates the proposed \emph{Anatomically-conditioned Latent Diffusion Model (ALDM)}, a novel framework for three-dimensional (3D) volumetric glioma MRI synthesis across heterogeneous source and target glioma populations in few-shot learning scenarios. ALDM addresses a significant research gap by integrating latent diffusion processes with anatomical structure conditioning to achieve data-efficient, spatially coherent synthesis optimized for downstream glioma classification tasks. By synthesizing high-quality target-domain volumes from limited source-domain data, ALDM enables practical data augmentation for glioma datasets while preserving the domain-specific imaging characteristics essential for classification accuracy.
The main contributions are summarized as follows:
\begin{enumerate}
    \item We propose ALDM, which combines a 3D variational autoencoder (VAE) for source-domain anatomical representation learning with a conditional LDM guided by tumor masks, enabling controlled synthesis of multimodal volumetric glioma MRI across heterogeneous source and target domain imaging protocols, generating anatomically coherent T1-weighted (T1), T2-weighted (T2), and T2-weighted fluid-attenuated inversion recovery (FLAIR) images with preserved tumor heterogeneity and domain-specific imaging characteristics.
    \item We show ALDM synthesizes anatomically consistent 3D MRI volumes when transferring from a data-rich and homogeneous glioblastoma (GBM) source domain \cite{bakas2021multi} to a data-scarce and heterogeneous pre-operative diffuse glioma (PDGM) target domain \cite{calabrese2022university} under few-shot and consistently outperform GAN-based and hybrid baselines in both image-level similarity metrics and downstream classification performance.
\end{enumerate}

%% file: sec/2_rw.tex
\section{Related Work}
\subsection{Domain Heterogeneity in Glioma MRI Datasets}
Domain heterogeneity in glioma MRI arises from both biological variability and acquisition differences across sites. Public datasets such as BraTS and TCGA/TCIA aggregate scans acquired with different scanner vendors, field strengths, coils, reconstruction pipelines, and protocol settings, which introduce systematic shifts in resolution, noise, and contrast even after curation and preprocessing \cite{menze2014multimodal,bakas2017advancing}.

These site effects create a distribution mismatch between development and deployment data and can substantially degrade out-of-domain performance \cite{abbad2025exploring}. In glioma segmentation, models that perform well on curated challenge data often drop in accuracy on routine clinical MRI, where image quality is more variable, and modality sets may be incomplete. External multi-center studies report clear gaps between BraTS-style evaluation and real hospital data, and suggest improved robustness when training includes more heterogeneous cohorts or explicitly handles missing modalities \cite{pemberton2023multi}. More generally, deep models may rely on scanner- or protocol-specific cues correlated with labels in the training set, reducing robustness under shifts in hardware, pulse sequences, or preprocessing \cite{bento2022deep}.

To improve cross-site robustness, common approaches include statistical harmonization that models sites as batch effects \cite{fortin2018harmonization}, domain adaptation that encourages domain-invariant representations \cite{kamnitsas2017unsupervised}, and generative harmonization or synthesis for scanner-style mapping and missing-sequence completion \cite{roca2025iguane}. Despite progress, reliable generalization without target labels or site-specific retraining remains difficult, especially when protocols and patient populations differ substantially across institutions.

\subsection{Generative Models for Volumetric Glioma MRI Synthesis}
Data scarcity and class imbalance in brain tumor cohorts have driven the development of generative models for MRI augmentation. While early GAN-based methods primarily synthesized 2D slices, recent approaches such as AGGrGAN \cite{mukherkjee2022brain} and LASTGAN \cite{na2025laplacian} have improved fidelity and domain specificity by incorporating multi-component aggregation and style encoders.
Generating full 3D tumor volumes is more challenging because models must preserve spatial consistency across depth. Recent work has therefore moved toward volumetric synthesis. A 3D VQ-GAN combined with a Transformer has been proposed to generate high-resolution tumor ROIs within MRI volumes \cite{zhou2025generating}. Diffusion models have also shown strong fidelity and diversity for 3D medical synthesis. Med-DDPM uses semantic conditioning within a 3D diffusion framework and achieves segmentation performance close to that on real data when synthetic volumes are used for augmentation \cite{dorjsembe2024conditional}. Similarly, mask-conditioned latent diffusion with a 3D autoencoder enables multi-contrast tumor volume generation from tumor masks while maintaining anatomical consistency \cite{truong2024synthesizing}.

Despite these advances, data-efficient 3D glioma MRI synthesis remains insufficiently studied. Most 3D GAN or diffusion pipelines still rely on moderate training set sizes, and few-shot settings with strong cross-domain shift have not been systematically validated. This gap motivates the development of generative frameworks that can learn from very limited target data while producing anatomically plausible and structurally consistent 3D MRI tumor volumes. 

%% file: sec/3_method.tex
\section{Architecture}
\begin{figure}[t]
    \centering
    \includegraphics[width=\linewidth]{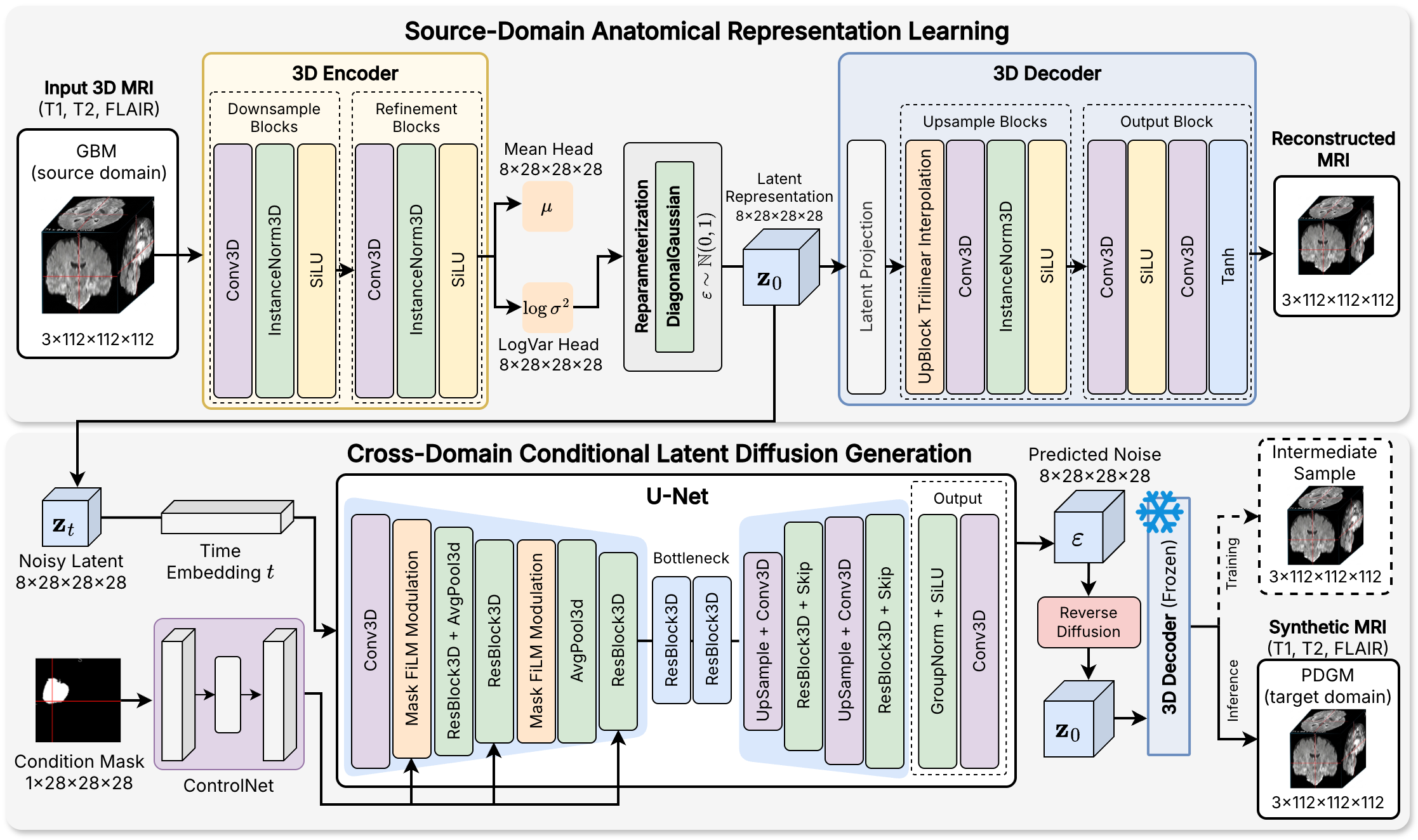}
    \caption{Overview of the proposed Anatomically-conditioned Latent Diffusion Model (ALDM) for cross-domain MRI synthesis. A VAE encoder compresses 3D GBM MRI volumes into a latent space, where a U-Net-based denoising diffusion model iteratively denoises under anatomical-mask conditioning. The decoded output produces synthetic PDGM MRI volumes.}
    \label{fig:arch}
\end{figure}

The proposed ALDM is a conditional latent-diffusion-based framework designed for cross-domain synthesis of 3D MRI volumes under data scarcity. As illustrated in \autoref{fig:arch}, the architecture decomposes the generative process into two stages: (i) source-domain anatomical representation learning using a VAE, and (ii) cross-domain conditional latent diffusion generation in the learned latent space.

\subsection*{Problem Formulation}
Let the source (data-rich, homogeneous) domain be $\mathcal{D}_s=\{(x_i^{s}, m_i^{s})\}_{i=1}^{N_s}$ and the target (data-scarce, heterogeneous) domain be $\mathcal{D}_t=\{(x_j^{t}, m_j^{t})\}_{j=1}^{N_t}$, where $x \in \mathbb{R}^{C \times D \times H \times W}$ denotes a multi-modal 3D MRI volume and $m$ is the corresponding tumor segmentation mask.
Here, $C=3$ corresponds to T1, T2, and FLAIR, and $(D,H,W)$ are the volumetric spatial dimensions.
Our goal is to learn a conditional generative model for the target domain that synthesizes a realistic volume $\hat{x}^{t}$ given an anatomical prior derived from the mask.
We denote this prior as $c=f(m)$, where $f(\cdot)$ may include the binary mask and auxiliary mask-derived cues (e.g., edges or distance transforms).
Generation is performed from a latent variable $z$ such that: $\hat{x}^{t} = G(z, c).$

We consider the \emph{few-shot} setting where $N_t = K \ll N_s$, with $K$ labeled target-domain examples available for adaptation.
The challenge is to preserve shared organ-level anatomy across domains while allowing controlled variation in tumor appearance guided by $c$.

\subsection{Source-Domain Anatomical Representation Learning}

In the first stage, a 3D VAE is trained on the data-rich GBM MRI dataset to learn a compact and anatomically consistent latent representation of volumetric brain images. Given an input MRI volume $\mathbf{x} \in \mathbb{R}^{C \times D \times H \times W}$, where $C=3$ corresponds to the T1, T2, and FLAIR modalities and $(D,H,W)=(112,112,112)$ after preprocessing, the VAE encoder $q_\phi(\mathbf{z}\mid\mathbf{x})$ compresses the high-resolution input into a lower-dimensional latent tensor $\mathbf{z} \in \mathbb{R}^{z_C \times d \times h \times w}$ with $z_C=8$ and $(d,h,w)=(28,28,28)$. This compression significantly reduces spatial complexity while preserving global anatomical structure, making subsequent generative modeling more stable and memory efficient.

The encoder is implemented as a stack of 3D convolutional blocks with kernel size $3\times3\times3$. Two strided convolution layers perform spatial downsampling by a factor of $4$, yielding successive spatial resolutions of $112$, $56$, and $28$. At each stage, instance normalization and nonlinear activations are applied to ensure stable training under the small batch sizes typical of volumetric medical imaging. Additional convolutional refinement blocks are inserted at the latent resolution to increase representational capacity without further reducing spatial detail, allowing the encoder to capture fine-grained anatomical variations. The encoder predicts the parameters of a diagonal Gaussian posterior distribution as:
\begin{equation}
   q_\phi(\mathbf{z} \mid \mathbf{x}) =\mathcal{N}\left(\boldsymbol{\mu}_\phi(\mathbf{x}),\operatorname{diag}(\boldsymbol{\sigma}_\phi^2(\mathbf{x}))\right), 
\end{equation}
from which latent samples are drawn using the reparameterization. To prevent numerical instability, the predicted log-variance is clamped to a fixed range during training.

The decoder mirrors the encoder architecture and reconstructs volumetric MRI data from latent samples. It consists of two successive upsampling stages implemented using trilinear interpolation followed by 3D convolution, resulting in spatial resolutions of $28$, $56$, and $112$. This design avoids checkerboard artifacts commonly associated with transposed convolutions and encourages smooth volumetric reconstructions. A final $\tanh$ activation constrains voxel intensities to the normalized range $[-1,1]$. By reconstructing full 3D volumes rather than independent slices, the VAE enforces volumetric coherence across slices and preserves consistent anatomical structure throughout the brain.
The VAE is trained using a weighted combination of reconstruction and Kullback–Leibler (KL) divergence losses as:
\begin{equation}
\mathcal{L}_{\text{VAE}} =
\lambda_{\text{rec}} \, \|\mathbf{x} - \hat{\mathbf{x}}\|_1
+ \lambda_{\text{KL}}, \quad
\mathrm{KL}\big(q_\phi(\mathbf{z}\mid\mathbf{x}) \,\|\, \mathcal{N}(\mathbf{0},\mathbf{I})\big).
\end{equation}
where the reconstruction term encourages voxel-level fidelity and the KL term regularizes the latent distribution toward a unit Gaussian prior. A deliberately small $\lambda_{\text{KL}}$ is used to mitigate posterior collapse and retain anatomically meaningful variability in the latent space. 
To prevent posterior collapse during early training, the KL warm-up is applied. The KL weight is linearly increased over a fraction $\rho = 0.35$ of the total training steps:
\begin{equation}
    \lambda_{\text{KL}}(t) =
    \begin{cases}
    \lambda_{\text{KL}} \cdot \frac{t}{\rho T}, & t \leq \rho T, \\
    \lambda_{\text{KL}}, & t > \rho T,
    \end{cases}
\end{equation}
where $t$ denotes the current training step and $T$ the total number of steps.

A 3D gradient consistency loss further enforces structural fidelity and encourages sharp anatomical boundaries:
\begin{equation}
    \mathcal{L}_{\nabla} =
    \sum_{a \in \{x,y,z\}}
    \left\| \nabla_a \hat{\mathbf{x}} - \nabla_a \mathbf{x} \right\|_1.
\end{equation}

After training, the encoder and decoder weights are frozen, and the encoder is used exclusively to extract latent representations for diffusion-based generation.

\subsection{Cross-Domain Conditional Latent Diffusion Generation}
In the second stage, a conditional U-Net-based Denoising Diffusion Probabilistic Model (DDPM) is trained in the VAE's latent space. Instead of operating on full-resolution voxel grids, diffusion is performed on normalized latent tensors, which significantly reduces computational cost while preserving global anatomical structure. This design is particularly important for 3D MRI data, where direct diffusion in voxel space is prohibitively expensive. The latent diffusion model is trained to transform Gaussian noise into anatomically plausible latent representations, which are subsequently decoded into synthetic MRI volumes.

\subsubsection{Forward Diffusion in Latent Space}
Forward diffusion process gradually corrupts a clean latent representation $\mathbf{z}_0$ by injecting Gaussian noise over $T$ timesteps.
At each timestep $t$, the noisy latent $\mathbf{z}_t$ is obtained as:
\begin{equation}
    \mathbf{z}_t =
    \sqrt{\bar{\alpha}_t}\,\mathbf{z}_0 +
    \sqrt{1 - \bar{\alpha}_t}\,\boldsymbol{\epsilon},
    \quad
    \boldsymbol{\epsilon} \sim \mathcal{N}(\mathbf{0}, \mathbf{I}),
\end{equation}
where $\bar{\alpha}_t = \prod_{s=1}^{t}(1 - \beta_s)$, and $\{\beta_t\}_{t=1}^{T}$ follows a linear schedule from $10^{-4}$ to $2 \times 10^{-2}$
over $T=1000$ timesteps. This process progressively removes structural information from the latent representation, yielding a sequence of increasingly noisy latents that serve as training inputs for the reverse denoising model. Operating in the VAE latent space ensures that the forward diffusion primarily perturbs semantically meaningful anatomical features rather than low-level voxel noise.

\subsubsection{Reverse Denoising with a Conditional 3D U-Net}
The reverse diffusion process is parameterized by a 3D U-Net, denoted as $\boldsymbol{\epsilon}_\theta(\mathbf{z}_t, t, \mathbf{c})$, which is trained to predict the noise injected at each timestep. 
The U-Net follows an encoder–decoder architecture with skip connections and processes latent tensors at spatial resolutions of $28$, $14$, and $7$, before symmetrically restoring the resolution to $28$.
Each resolution level consists of residual 3D convolutional blocks with group normalization, which provides stable training under the small batch sizes imposed by volumetric MRI data.

Temporal information is incorporated via sinusoidal timestep embeddings as:
\begin{equation}
    \gamma(t) =
    \left[
    \cos(\omega_k t), \sin(\omega_k t)
    \right]_{k=1}^{d/2},
\end{equation}
which are projected through multilayer perceptrons and injected into each residual block as feature-wise biases. 
This conditioning allows the network to adapt its denoising behavior across diffusion steps and effectively model the time-dependent structure of the reverse process. 
By predicting the noise component, the model employs standard DDPM formulation and benefits from improved training stability.

\subsubsection{Conditional Generation via Latent Diffusion}
To enable cross-domain adaptation and explicit anatomical guidance, the latent diffusion model is conditioned on tumor segmentation masks derived from both the source and target domains. Conditioning is incorporated through complementary mechanisms operating at multiple spatial scales. 

First, a lightweight FiLM-style modulation applies mask-dependent affine transformations to intermediate feature maps, providing a global bias toward tumor-aware representations without significantly increasing model complexity as follows:
\begin{equation}
    \mathbf{h}' =
    \mathbf{h} \odot (1 + 0.1\,\gamma) + 0.1\,\beta,
\end{equation}
where $(\gamma,\beta)$ are predicted from global mask statistics.

Second, additional anatomical control signals, including edge maps and soft distance transforms derived from the tumor mask, are processed by a ControlNet \cite{zhang2023adding}. The resulting residual feature maps are injected into the main U-Net at multiple resolutions, enforcing spatial alignment between generated latent structures and anatomical priors.
This multi-scale conditioning strategy encourages the preservation of tumor location, shape, and spatial extent during generation, while still allowing the diffusion process to model realistic intensity variation and global brain anatomy.
By conditioning on anatomical structure rather than domain-specific intensity statistics alone, the model effectively transfers structural knowledge from the data-rich GBM domain to the data-scarce PDGM domain.
The diffusion model is trained by minimizing the mean-squared error between the true injected noise and the predicted noise with additional spatial weighting applied within tumor regions to prevent lesion attenuation during denoising. 

\subsubsection{Conditional Guidance and Loss Weighting}
To enable flexible control over anatomical conditioning strength, we adopt classifier-free guidance during latent diffusion training.
Specifically, conditioning inputs are randomly dropped with probability
$p_{\text{drop}} = 0.1$, such that the denoising network learns both conditional and unconditional noise predictions.
During inference, the guided noise estimate is computed as:
\begin{equation}
\boldsymbol{\epsilon}_{\text{guided}} =
\boldsymbol{\epsilon}_\theta(\mathbf{z}_t, t, \varnothing)
+ s \left(
\boldsymbol{\epsilon}_\theta(\mathbf{z}_t, t, \mathbf{c})
- \boldsymbol{\epsilon}_\theta(\mathbf{z}_t, t, \varnothing)
\right),
\end{equation}
where $\varnothing$ denotes null conditioning and $s$ is the guidance scale. This formulation allows explicit trade-off between anatomical adherence and sample diversity at generation time.

To further preserve tumor structure during denoising, the diffusion loss is spatially weighted using the ground-truth tumor mask.
In addition to the binary tumor region, a dilated neighborhood is included to encourage smooth structural transitions at lesion boundaries. The dilation is implemented using a $3 \times 3 \times 3$ structuring element in latent space.
Formally, the weighted diffusion objective is expressed as
\begin{equation}
\mathcal{L}_{\text{diff}} =
\mathbb{E}_{t,\boldsymbol{\epsilon}}
\left[
w(\mathbf{m})
\left\|
\boldsymbol{\epsilon} -
\boldsymbol{\epsilon}_\theta(\mathbf{z}_t, t, \mathbf{c})
\right\|_2^2
\right],
\end{equation}
where $w(\mathbf{m})$ assigns higher weights to voxels within the tumor and its local neighborhood.
The overall influence of structural control signals is modulated by a global scaling factor $\lambda_{\text{ctrl}} = 1.0$, which is applied uniformly across all conditioning pathways.

After the reverse diffusion process converges, the denoised latent representation $\mathbf{z}_0$ is passed through the frozen VAE decoder to produce a synthetic 3D MRI volume as:
\begin{equation}
    \mathbf{x}_{\text{syn}} = \mathrm{Dec}(\mathbf{z}_0), \quad \mathbf{z}_0 = \mathrm{DDPM}^{-1}(\mathcal{N}(0,I), \mathbf{c}).
\end{equation}
This formulation tightly couples anatomical conditioning with diffusion-based generation, enabling anatomically coherent, controllable, and data-efficient cross-domain MRI synthesis under both few-shot and zero-shot target-domain settings.

%% file: sec/4_exp.tex
\section{Experiment Setup}
\subsection{Dataset}
Our experiment uses a GBM dataset \cite{bakas2021multi} and a PDGM dataset \cite{calabrese2022university}, both with T1, T2, and FLAIR modalities, stored primarily in NIfTI format. The GBM dataset contains approximately 828{,}000 image slices derived from 3D MRI volumes of multiple subjects. In addition to imaging, the GBM dataset includes structured clinical metadata files with data types including MR, molecular test results, and demographic records. While PDGM consists of approximately 12{,}000 images, with accompanying metadata categories including MR, measurement, demographic, follow-up, and diagnosis. This strong imbalance between a large source-domain cohort (GBM) and a limited target-domain cohort (PDGM) makes the setting well-suited for evaluating few-shot cross-domain generative transfer.
Separately, the full PDGM dataset is used to train the downstream CNN classifier for evaluation, ensuring the classifier learns target-domain decision boundaries from real target data rather than synthetic outputs.

% In our experiments, the PDGM data is intentionally constrained to a few-shot subset of 16 images for target-domain adaptation and evaluation, matching the intended low-data regime. 

\subsection{Baselines} 
We compare our proposed ALDM performance against our three re-implemented representative generative baselines commonly used for medical image synthesis and cross-domain adaptation. These baselines span adversarial, multi-discriminator, and hybrid latent–adversarial paradigms, providing a comprehensive comparison across modeling strategies. (1) conditional GAN (CGAN) \cite{mirza2014conditional}, which conditions the generator on auxiliary information to guide image synthesis across domains. 
% This model serves as a standard adversarial baseline for conditional image generation; 
(2) 3M-CGAN extension by employing an ensemble of three discriminators, each operating on three corresponding MRI modalities \cite{xin2020multi}. This multi-discriminator design encourages modality-specific realism and has been shown to improve training stability and visual fidelity in multi-channel medical imaging tasks.
(3) Hybrid VAE-GAN model \cite{larsen2016autoencoding}, which combines a VAE for latent-space representation learning with a conditional adversarial objective to enhance sample realism. 
% By jointly optimizing reconstruction and adversarial losses, this model aims to balance diversity and fidelity in generated MRI volumes.

\subsection{Evaluation Metrics} 
In our experiment, we conducted the performance comparison across both the data-rich GBM domain and the data-scarce PDGM domain through two standards:
\begin{enumerate}
    \item Image-Level Fidelity: We compute the following metrics at the patient-wise fidelity level to preserve volumetric integrity and ensure clinically meaningful evaluation that reflects real-world diagnostic workflows:
    \begin{enumerate}
        \item \textit{Fréchet Inception Distance (FID)} measures distributional similarity between real and synthesized MRI volumes, with lower values indicating closer alignment with the target distribution.
        \item \textit{The Structural Similarity Index Measure (SSIM)} quantifies local and global structural correspondence between generated and real MRI volumes, with higher values indicating greater anatomical consistency.
    \end{enumerate}
    Specifically, we evaluate all models on 64 target-domain subjects, each consisting of 112 aligned axial slices per modality.
    For each subject, SSIM and FID are computed by aggregating metrics across the 112 generated and real slices, yielding a single fidelity score per patient.
    Final patient-wise results are reported as the mean across the 64 subjects.
 
    \item Downstream classification: To assess the diagnostic utility of the generated images, a downstream AlexNet-based classifier \cite{krizhevsky2012imagenet} was evaluated on synthetic MRI data after being trained exclusively on real samples from the original dataset. This evaluation framework was used to ensure the trained CNN was adhered to classify the images with high accuracy, where downstream performance serves as a proxy for clinical usefulness. Performance is measured using balanced classification accuracy (BAcc), F1 score, and the area under the ROC curve (AUC).
\end{enumerate}

%% file: sec/5_res.tex
\section{Results}
\subsection{Quantitative Evaluation}
The quantitative results, summarized in \autoref{tab:comparative_metrics} and visualized in \autoref{fig:radar_comparison}, highlight the effectiveness of the ALDM framework in the data-scarce PDGM target domain.

\begin{table}[b]
\captionsetup{margin=20pt}
\centering
\caption{Comparative performance of generative models on the target PDGM domain. The best is in \textbf{bold}, the second best is \underline{underlined}. $\uparrow$/$\downarrow$ indicate the higher/lower, the better performance. Metrics are reported as mean values aggregated across 3$\times$5-fold cross-validation.}
\begin{tabular}{l*{4}{>{\centering\arraybackslash}p{1.4cm}}>{\centering\arraybackslash}p{2.5cm}}
\toprule
& \multicolumn{2}{c}{\textbf{(a) Fidelity}} & \multicolumn{3}{c}{\textbf{(b) Downstream Classification}} \\ 
\cmidrule(lr){2-3} \cmidrule(lr){4-6} 
\textbf{Model} & FID$\downarrow$ & SSIM$\uparrow$ & BAcc$\uparrow$ & F1$\uparrow$ & AUC$\uparrow$ \\
\midrule
CGAN \cite{mirza2014conditional} & 145.22 & 0.374 & 0.764 & 0.720 & 0.876 $\pm$ 0.142 \\
3M-CGAN \cite{xin2020multi} & 116.48 & 0.680  & 0.780 & 0.731 & 0.866 $\pm$ 0.004 \\
VAE-GAN \cite{larsen2016autoencoding}  & 88.18 & \textbf{0.750}  & 0.751 & 0.675 & 0.882 $\pm$ 0.004 \\
\midrule
ALDM ($K$=16, $s$=0.3) & 88.02 & \underline{0.716} & 0.780 & 0.730 & 0.871 $\pm$ 0.004   \\
ALDM ($K$=16, $s$=0.5)  & 88.08 & 0.714 & 0.774 & 0.721 &  0.877 $\pm$ 0.004  \\
ALDM ($K$=16, $s$=1.0)& \underline{87.52} & 0.715 & 0.783 & 0.733 & 0.897 $\pm$ 0.005  \\
ALDM ($K$=10, $s$=3.0) & 95.08 & 0.699 & \underline{0.856} & \underline{0.832} & \underline{0.948 $\pm$ 0.003}  \\
\midrule
\rowcolor{gray!10}
ALDM ($K$=16, $s$=3.0) & \textbf{85.40} & 0.712 & \textbf{0.875} & \textbf{0.836} & \textbf{0.987 $\pm$ 0.001} \\
\bottomrule
\end{tabular}
\label{tab:comparative_metrics}
\end{table}

\begin{figure}[b]
\centering
\captionsetup{margin=0pt}
\begin{minipage}[t]{0.48\textwidth}
\centering
\begin{tikzpicture}
\begin{polaraxis}[
    width=4.5cm,
    height=4.5cm,
    xtick={0,72,144,216,288},
    xticklabels={FID*, SSIM, BAcc, F1, AUC},
    xticklabel style={font=\tiny},
    ymin=0, ymax=1,
    ytick={0,0.5,1.0},
    yticklabel style={font=\tiny},
    legend style={at={(-0.3,0.5)}, anchor=east, font=\tiny, line width=0.5pt}, 
    grid=major,
]
\addplot[brown!50!orange!60, mark=*, mark size=1.5pt, line width=1pt] coordinates {
    (0,0) (72,0.374) (144,0.764) (216,0.720) (288,0.876)
} -- cycle;
\addlegendentry{CGAN}
\addplot[olive!60!yellow!70, mark=square*, mark size=1.5pt, line width=1pt] coordinates {
    (0,0.48) (72,0.680) (144,0.780) (216,0.731) (288,0.866)
} -- cycle;
\addlegendentry{3M-CGAN}
\addplot[teal!50!cyan!60, mark=triangle*, mark size=1.5pt, line width=1pt] coordinates {
    (0,0.96) (72,0.750) (144,0.751) (216,0.675) (288,0.882)
} -- cycle;
\addlegendentry{VAE-GAN}
\addplot[purple!60!magenta!70, mark=diamond*, mark size=2pt, line width=1.2pt] coordinates {
    (0,1.0) (72,0.712) (144,0.875) (216,0.836) (288,0.987)
} -- cycle;
\addlegendentry{ALDM}
\end{polaraxis}
\end{tikzpicture}
\caption{Multi-metric performance comparison (*FID normalized).}
\label{fig:radar_comparison}
\end{minipage}%
\hfill
\begin{minipage}[t]{0.48\textwidth}
\centering
\begin{tikzpicture}
\begin{axis}[
    width=5cm,
    height=4.5cm,
    xlabel={Scaling Factor ($s$)},
    ylabel={Performance Score},
    xlabel style={font=\tiny},
    ylabel style={font=\tiny},
    legend style={at={(-0.3,0.5)}, anchor=east, font=\tiny, line width=0.5pt}, 
    tick label style={font=\tiny},
    grid=major,
    ymin=0.65, ymax=1.0,
    xmode=log,
    log basis x=10,
    xtick={0.3,0.5,1,3},
    xticklabels={0.3,0.5,1.0,3.0},
]
\addplot[thick, cyan!70!blue, mark=*, mark size=2pt] coordinates {
    (0.3,0.780) (0.5,0.774) (1.0,0.783) (3.0,0.875)
};
\addlegendentry{BAcc}
\addplot[thick, magenta!80!red, mark=square*, mark size=2pt] coordinates {
    (0.3,0.730) (0.5,0.721) (1.0,0.733) (3.0,0.836)
};
\addlegendentry{F1}
\addplot[thick, teal!60!cyan, mark=triangle*, mark size=2pt] coordinates {
    (0.3,0.871) (0.5,0.877) (1.0,0.897) (3.0,0.987)
};
\addlegendentry{AUC}
\addplot[thick, violet!70!magenta, mark=diamond*, mark size=2pt] coordinates {
    (0.3,0.716) (0.5,0.714) (1.0,0.715) (3.0,0.712)
};
\addlegendentry{SSIM}
\end{axis}
\end{tikzpicture}
\caption{ALDM hyperparameter sensitivity analysis. Performance trends across different scaling factors ($s$) with fixed $K$=16.}
\label{fig:aldm_scaling}
\end{minipage}
\end{figure}

\subsubsection{Image-Level Fidelity}
Regarding image-level fidelity, the proposed ALDM ($K=16, s=3.0$) achieves the lowest FID of 85.40, indicating the highest distributional similarity to the real MRI volumes. While the VAE-GAN baseline achieves a higher SSIM of 0.750, it produces a higher FID (88.18) than the ALDM variants. The results suggest that while VAE-GAN excels at local pixel-wise reconstruction, the ALDM framework better captures the global distribution and stylistic nuances of the target domain. Other baselines, such as CGAN and 3M-CGAN, trail significantly in FID (145.22 and 116.48, respectively), underscoring the limitations of standard adversarial approaches in few-shot medical imaging contexts.

\subsubsection{Downstream Classification Evaluation}
The clinical utility of the synthesized images was assessed by training a downstream classifier on the real 12k image dataset and testing it on synthetic PDGM samples generated by different models. ALDM ($K=16, s=3.0$) outperformed all other models across the primary classification metrics, reaching a balanced accuracy (BAcc) of 0.875, F1 of 0.836, and a superior AUC of $0.987 \pm 0.001$.
In contrast, the VAE-GAN, despite its high SSIM, yielded the lowest F1 score (0.675) and the lowest BAcc (0.751) among the tested models, suggesting that its generated features may not translate as effectively into diagnostic tasks. The proposed model's consistent performance across BAcc, F1, and AUC confirms that the ALDM framework preserves more robust and discriminative features necessary for downstream medical applications.

\begin{figure}[b]
\centering
\begin{tabular}{>{\centering\arraybackslash}m{0.15\linewidth}cccc}
\textbf{Epoch} &
\textbf{FLAIR} &
\textbf{T1} &
\textbf{T2} &
\textbf{3D Image} \\ 
\midrule

Epoch 50 &
\includegraphics[valign=m,width=0.15\linewidth]{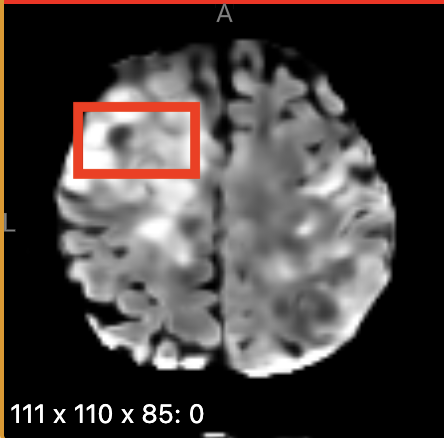} &
\includegraphics[valign=m,width=0.15\linewidth]{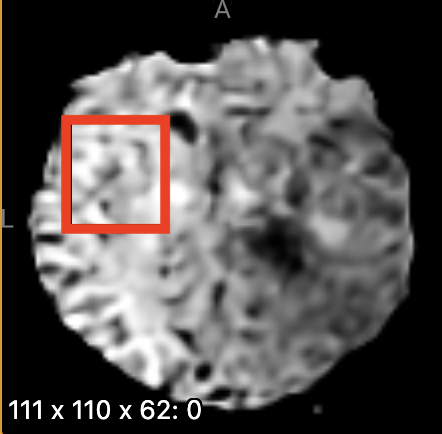} &
\includegraphics[valign=m,width=0.15\linewidth]{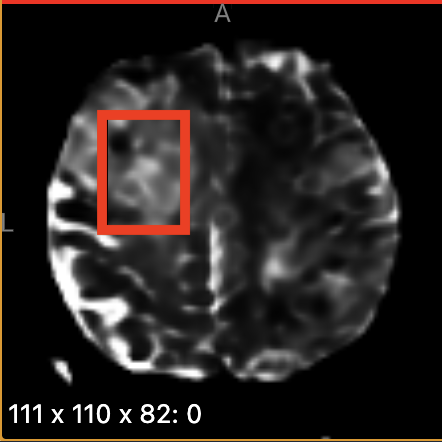} &
\includegraphics[valign=m,width=0.15\linewidth]{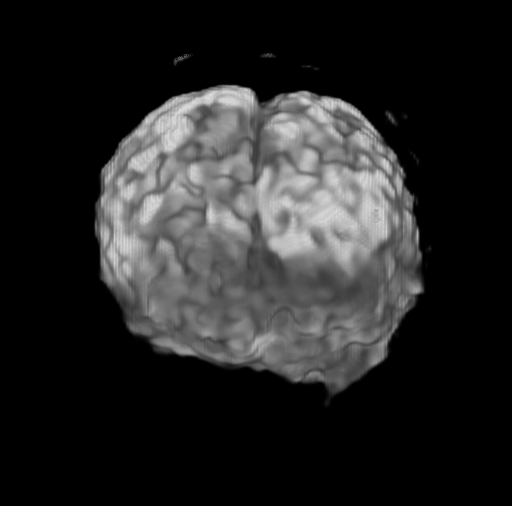} \\
\addlinespace[2pt]

Epoch 100 &
\includegraphics[valign=m,width=0.15\linewidth]{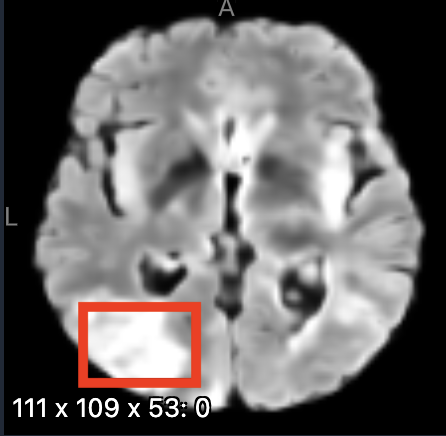} &
\includegraphics[valign=m,width=0.15\linewidth]{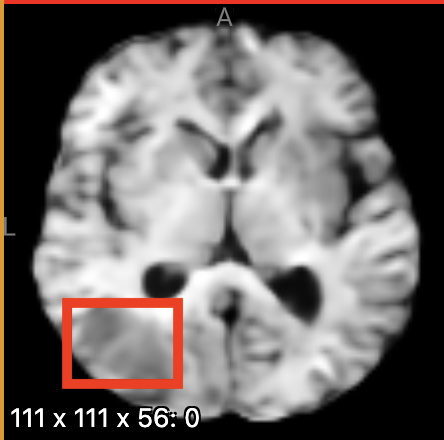} &
\includegraphics[valign=m,width=0.15\linewidth]{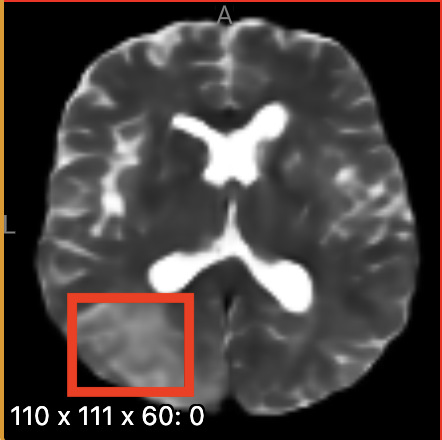} &
\includegraphics[valign=m,width=0.15\linewidth]{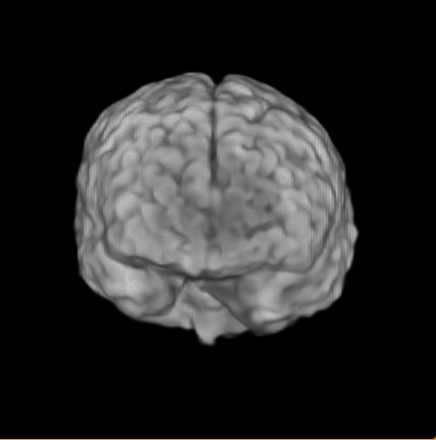} \\
\addlinespace[2pt]

Epoch 200 &
\includegraphics[valign=m,width=0.15\linewidth]{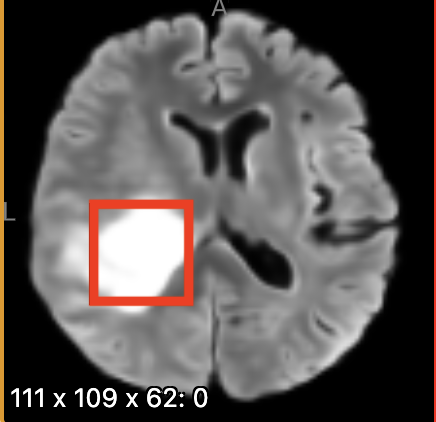} &
\includegraphics[valign=m,width=0.15\linewidth]{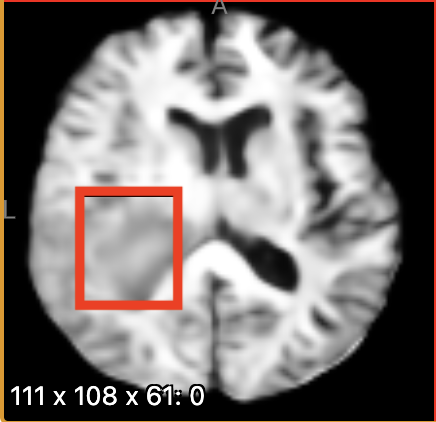} &
\includegraphics[valign=m,width=0.15\linewidth]{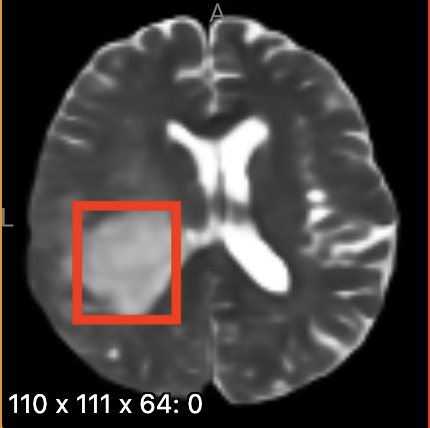} &
\includegraphics[valign=m,width=0.15\linewidth]{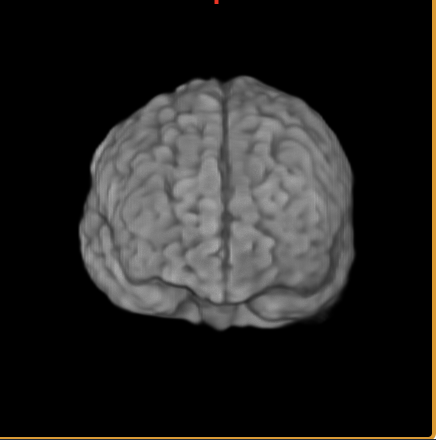} \\
\bottomrule
\end{tabular}

\caption{
Qualitative evolution of our proposed ALDM with synthesized MRI volumes over training epochs. Progressive improvements in anatomical fidelity, tumor delineation, and cross-modal consistency are observed as training converges.
}
\label{fig:epoch_progression}
\end{figure}

\begin{figure}[htbp]
\centering
\begin{tabular}{>{\centering\arraybackslash}m{0.2\linewidth}cccc}
\textbf{Model} &
\textbf{FLAIR} &
\textbf{T1} &
\textbf{T2} &
\textbf{3D Image} \\
\midrule
Ground Truth &
\includegraphics[valign=m,width=0.15\linewidth]{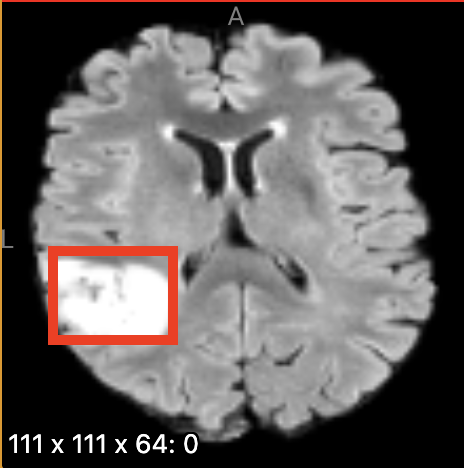} &
\includegraphics[valign=m,width=0.15\linewidth]{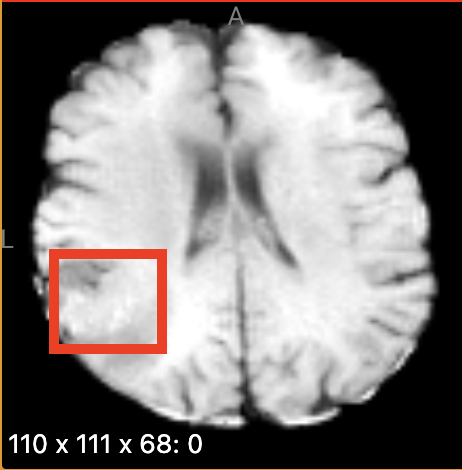} &
\includegraphics[valign=m,width=0.15\linewidth]{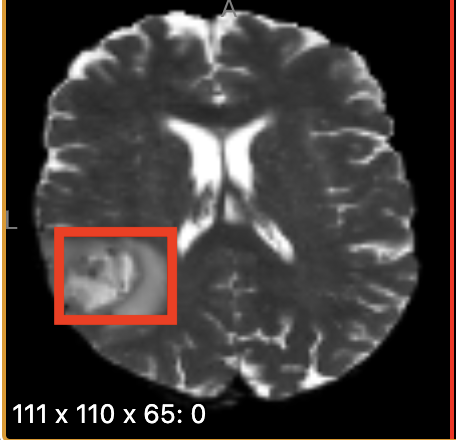} &
\includegraphics[valign=m,width=0.15\linewidth]{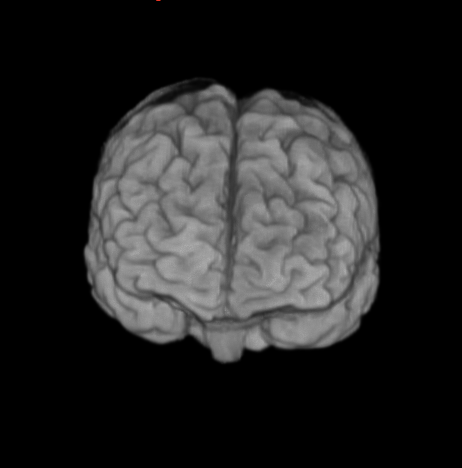} \\
\midrule
CGAN &
\includegraphics[valign=m,width=0.15\linewidth]{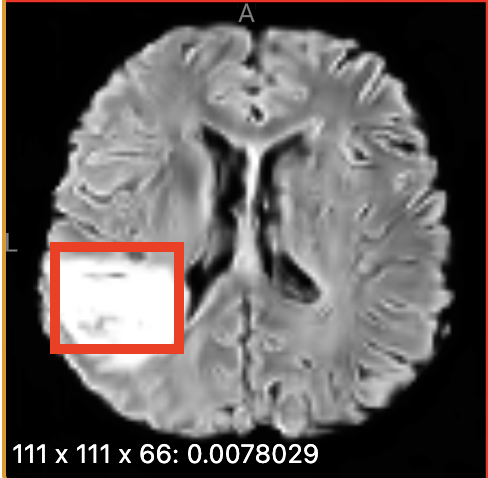} &
\includegraphics[valign=m,width=0.15\linewidth]{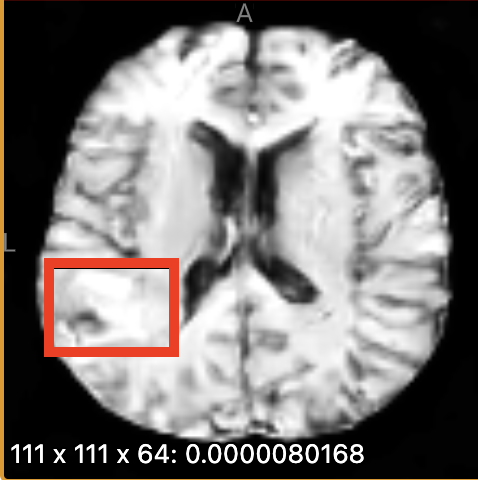} &
\includegraphics[valign=m,width=0.15\linewidth]{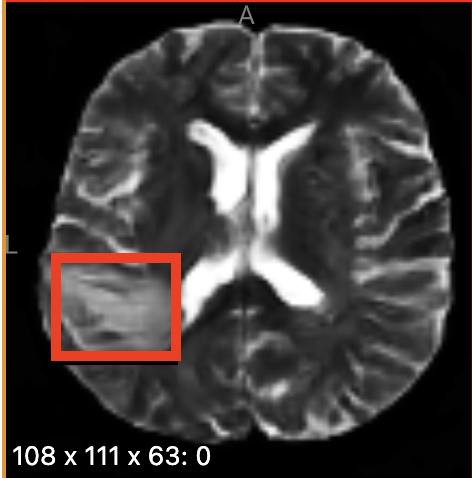} &
\includegraphics[valign=m,width=0.15\linewidth]{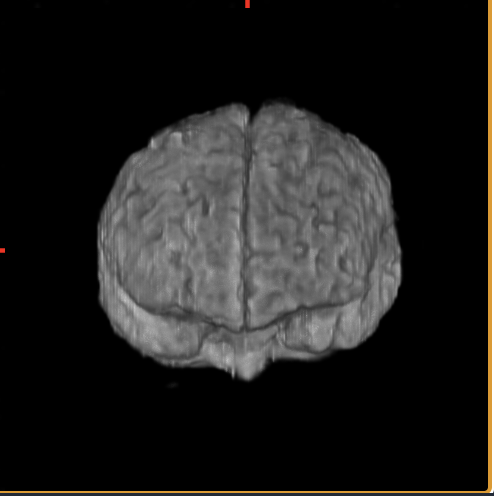} \\
\addlinespace[2pt]
3M-CGAN &
\includegraphics[valign=m,width=0.15\linewidth]{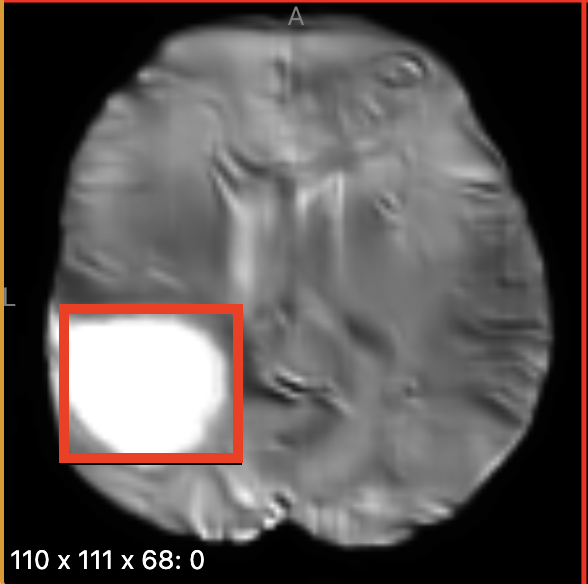} &
\includegraphics[valign=m,width=0.15\linewidth]{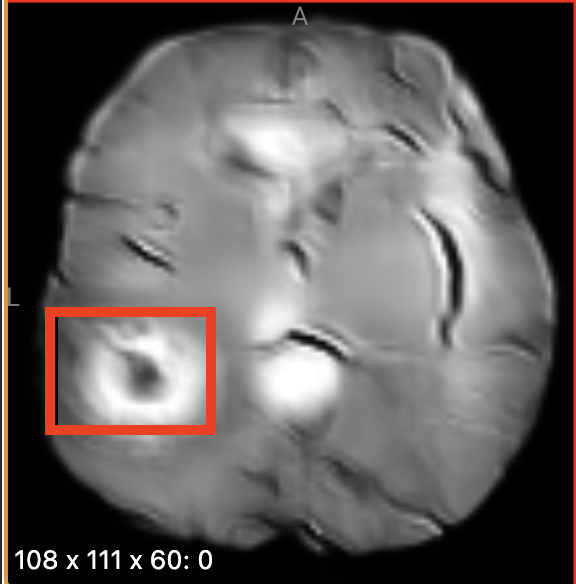} &
\includegraphics[valign=m,width=0.15\linewidth]{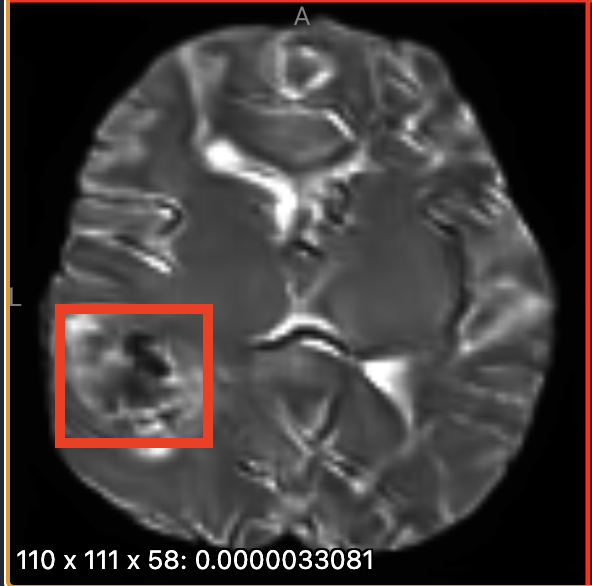} &
\includegraphics[valign=m,width=0.15\linewidth]{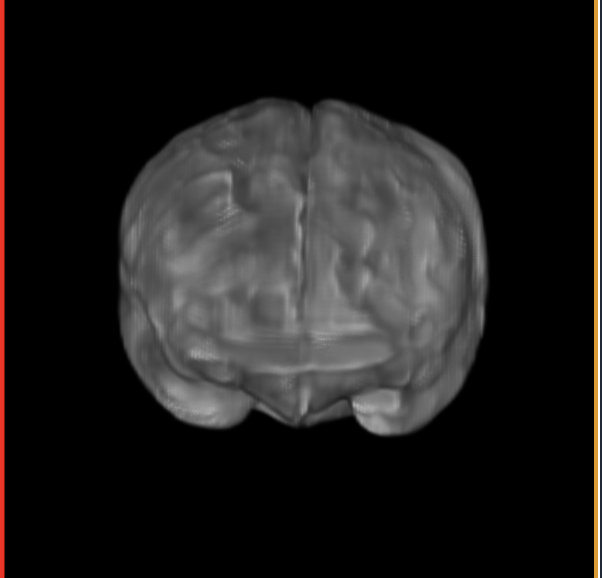} \\
\addlinespace[2pt]
VAE-GAN &
\includegraphics[valign=m,width=0.15\linewidth]{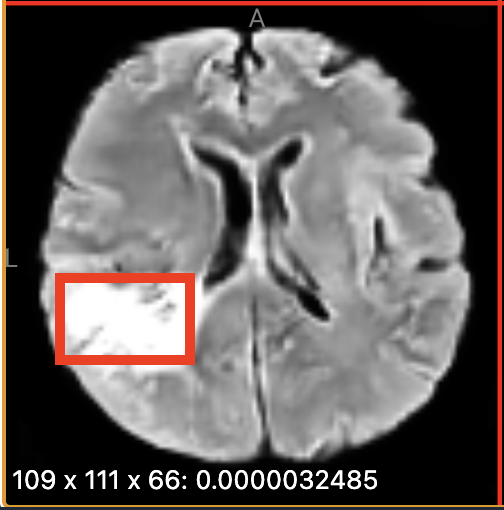} &
\includegraphics[valign=m,width=0.15\linewidth]{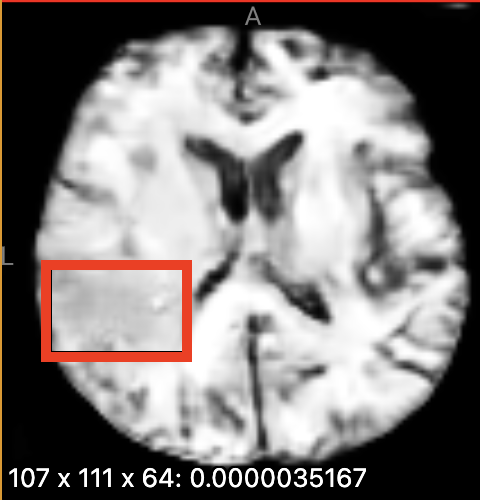} &
\includegraphics[valign=m,width=0.15\linewidth]{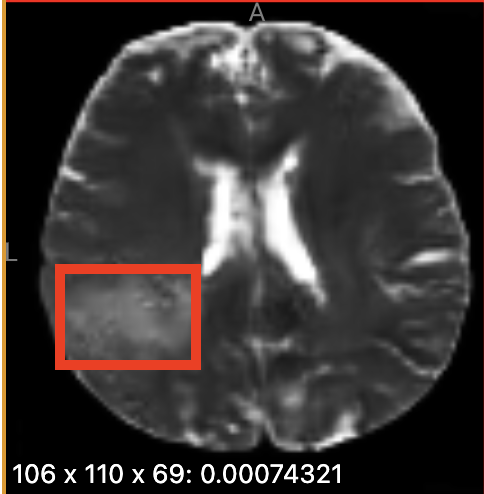} &
\includegraphics[valign=m,width=0.15\linewidth]{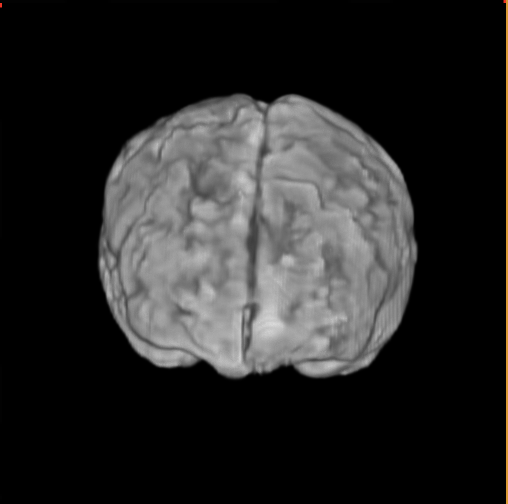} \\
\midrule
\makecell[c]{ALDM \\ ($K$=16, $s$=0.3)} & 
\includegraphics[valign=m,width=0.15\linewidth]{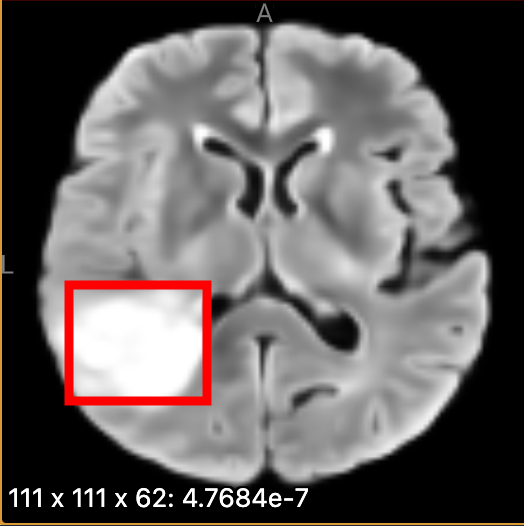} & 
\includegraphics[valign=m,width=0.15\linewidth]{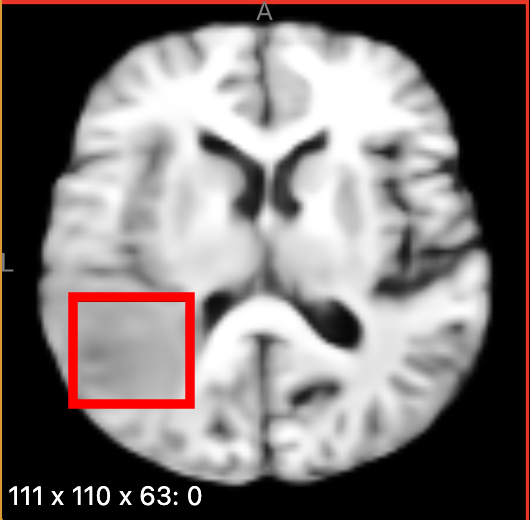} & 
\includegraphics[valign=m,width=0.15\linewidth]{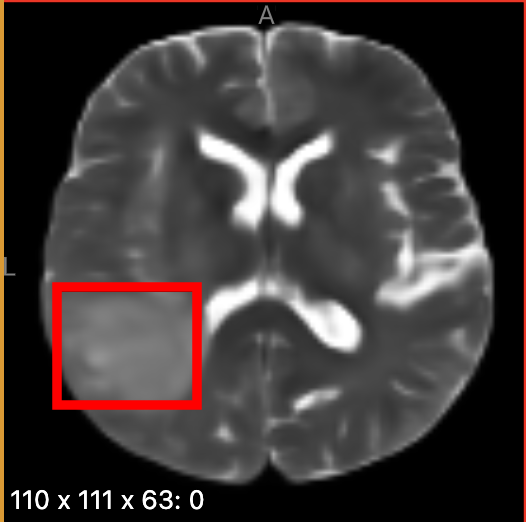} & 
\includegraphics[valign=m,width=0.15\linewidth]{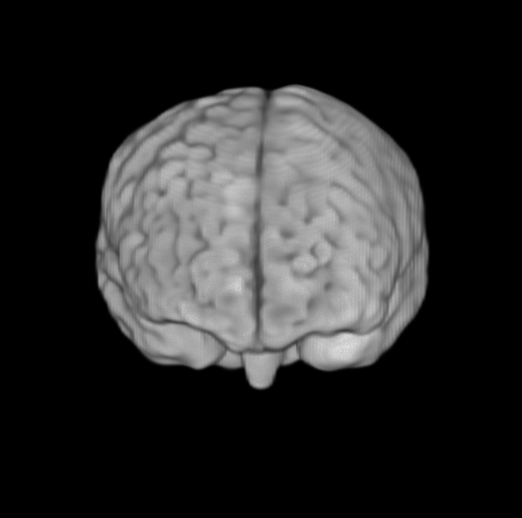} \\
\addlinespace[2pt]
\makecell[c]{ALDM \\ ($K=16, s=0.5$)} &
\includegraphics[valign=m,width=0.15\linewidth]{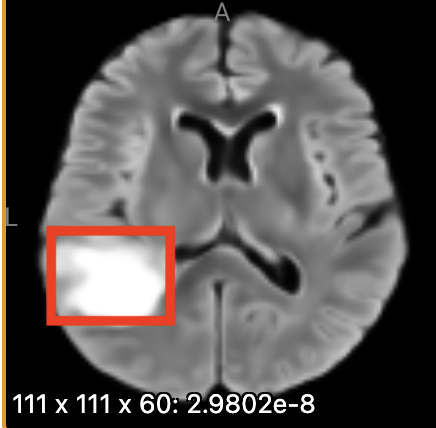} &
\includegraphics[valign=m,width=0.15\linewidth]{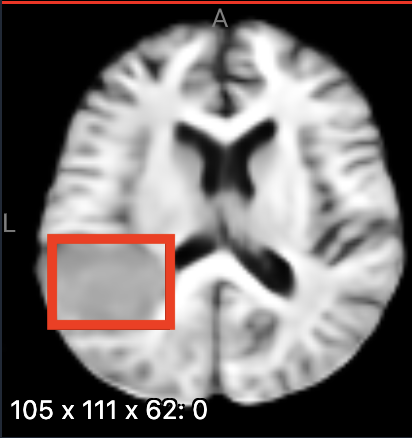} &
\includegraphics[valign=m,width=0.15\linewidth]{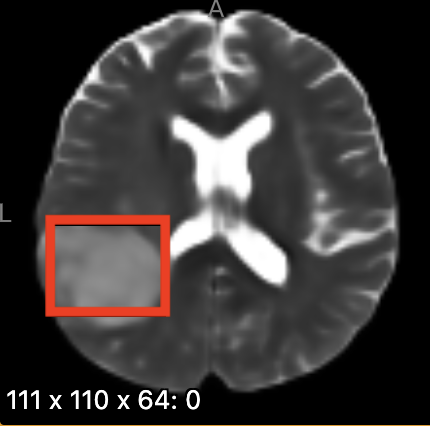} &
\includegraphics[valign=m,width=0.15\linewidth]{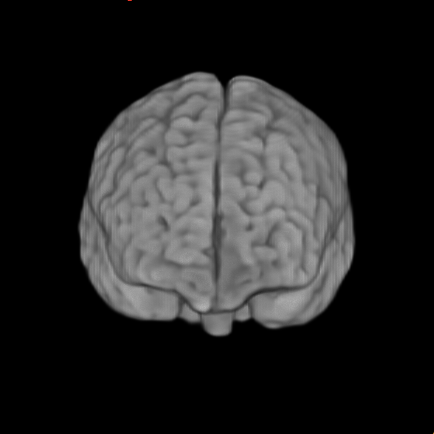} \\
\addlinespace[2pt]
\makecell[c]{ALDM \\  ($K$=16, $s$=1.0)} &
\includegraphics[valign=m,width=0.15\linewidth]{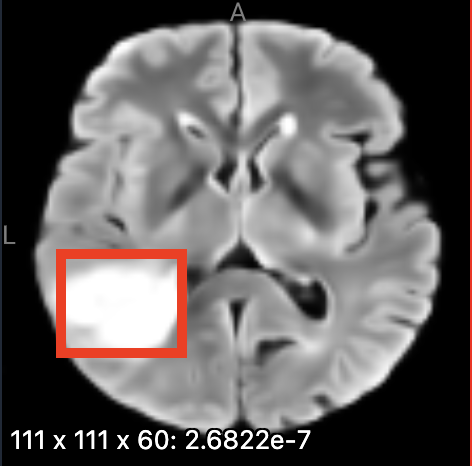} &
\includegraphics[valign=m,width=0.15\linewidth]{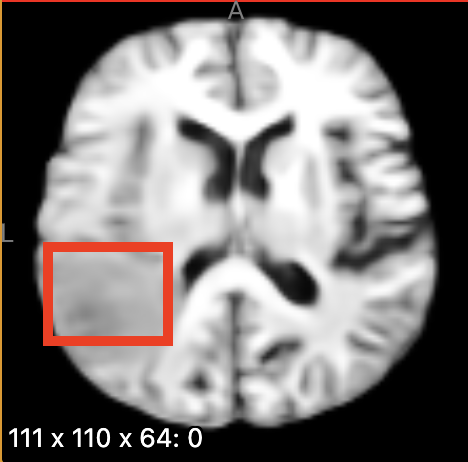} &
\includegraphics[valign=m,width=0.15\linewidth]{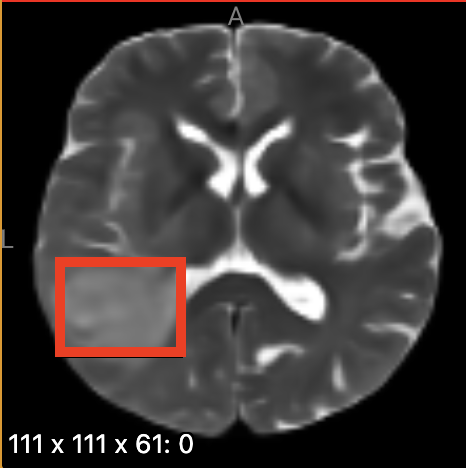} &
\includegraphics[valign=m,width=0.15\linewidth]{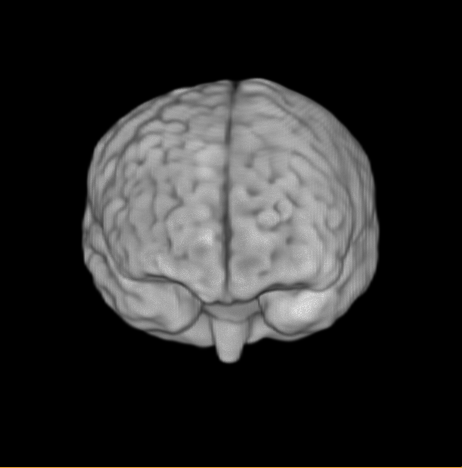} \\
\addlinespace[2pt]
\makecell[c]{ALDM \\  ($K$=10, $s$=3.0)} & 
\includegraphics[valign=m,width=0.15\linewidth]{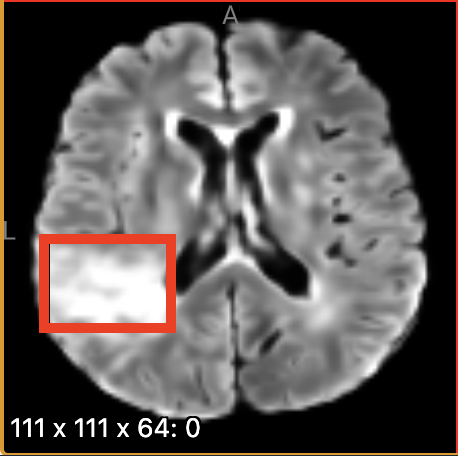} & 
\includegraphics[valign=m,width=0.15\linewidth]{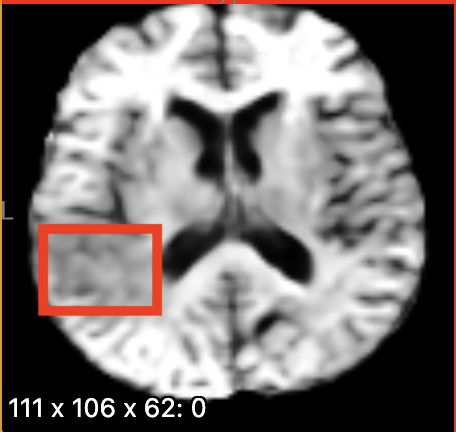} & 
\includegraphics[valign=m,width=0.15\linewidth]{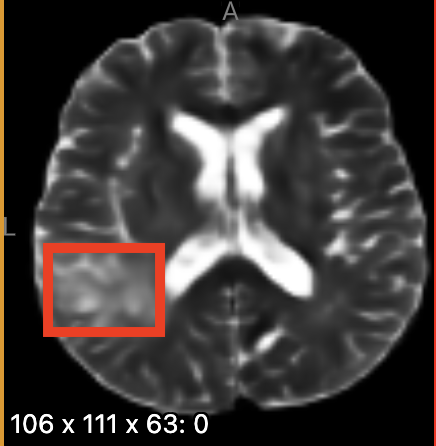} & 
\includegraphics[valign=m,width=0.15\linewidth]{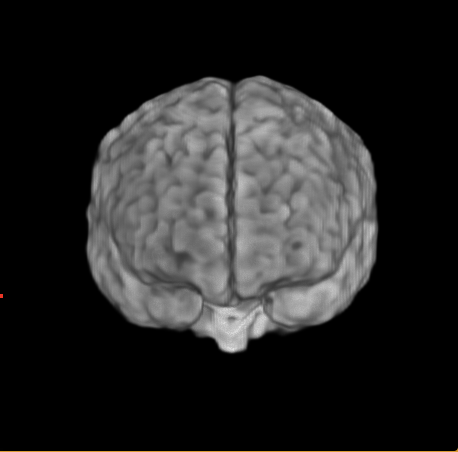} \\
\midrule
\makecell[c]{ALDM \\  ($K$=16, $s$=3.0)} & 
\includegraphics[valign=m,width=0.15\linewidth]{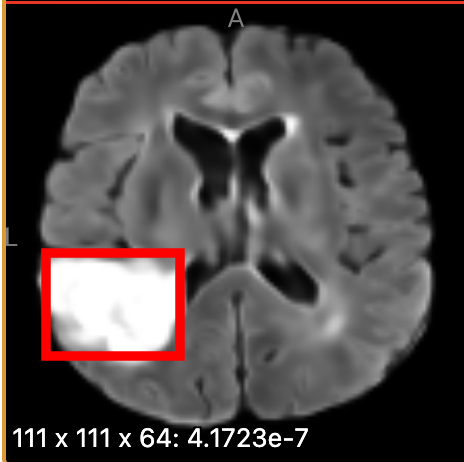} & 
\includegraphics[valign=m,width=0.15\linewidth]{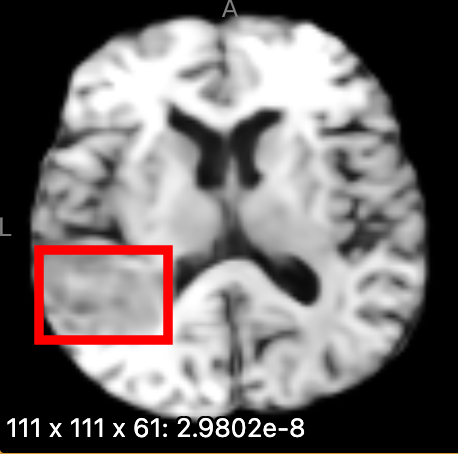} & 
\includegraphics[valign=m,width=0.15\linewidth]{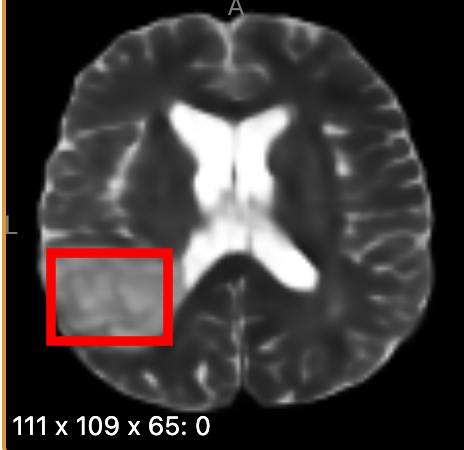} & 
\includegraphics[valign=m,width=0.15\linewidth]{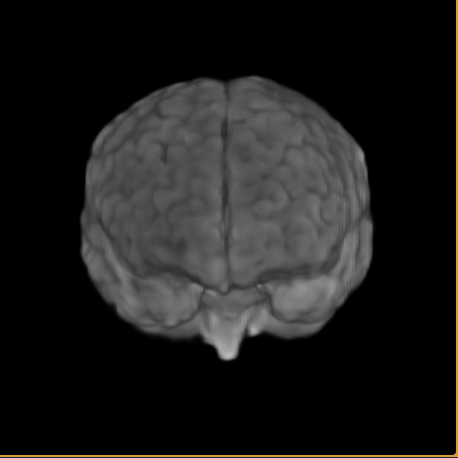} \\
\bottomrule
\end{tabular}
\caption{Qualitative comparison of synthesized MRI volumes across modalities. Tumor regions are highlighted in the red box to assess anatomical fidelity, cross-modal consistency, and 3D structural coherence.}
\label{fig:qualitative_grid}
\end{figure}

\subsubsection{Ablation Study}
\paragraph{\textbf{Impact of Scale Parameter ($s$)}} We examine the model's sensitivity to the classifier-free guidance scale parameter ($s$) at a fixed few-shot size of $K=16$. As shown in \autoref{tab:comparative_metrics} and \autoref{fig:aldm_scaling}, variations in the scale (from $s=0.3$ to $s=1.0$) result in relatively subtle performance shifts, with FID scores remaining stable around 88 and AUC values showing consistent improvement. While the default configuration of $s=3.0$ achieves the optimal balance, marked by a notable peak in AUC (0.987) and the lowest FID (85.40), the model’s performance is not overly sensitive to this choice. All tested scales consistently outperform the GAN and VAE-based baselines in downstream classification, indicating that the ALDM framework is robust to the specific tuning of the guidance scale.

\paragraph{\textbf{Impact of Few-Shot Examples ($K$)}}
The impact of the number of target-domain images ($K$) was also evaluated by comparing $K=10$ and $K=16$ at a scale of $s=3.0$. The results indicate that while increasing the sample size to $K=16$ provides the best overall results in both fidelity and classification accuracy, the difference between the two settings is marginal. Even with only 10 images, the ALDM achieves an AUC of 0.948 and a balanced accuracy of 0.856, which remain superior to those of CGAN, 3M-CGAN, and VAE-GAN. This suggests that while our chosen default of $K=16$ is optimal, the method remains highly effective in extremely low-data regimes, demonstrating that the specific choice of $K$ within this range is not critical for achieving competitive results.

This relative stability across different hyperparameter settings suggests that the ALDM framework is inherently robust. While fine-tuning $s$ and $K$ can provide incremental gains in fidelity and downstream accuracy, the overall success of the method is not overly dependent on a specific parameter choice. This flexibility is particularly advantageous in low-resource medical settings where exhaustive hyperparameter optimization may not be feasible.

\subsection{Qualitative Evaluation}
To visually validate the proposed framework's performance, we provide qualitative assessments of the training process and comparisons with baseline methods.

\subsubsection{Progressive Training Performance} 
As shown in \autoref{fig:epoch_progression}, the ALDM exhibits a clear progression in image quality as training epochs increase. At early stages (epoch 50), the model captures the general brain structure but produces blurry tumor regions with limited modality-specific contrast. By epoch 100, the anatomical fidelity improves significantly, with more distinct boundaries between white and gray matter. At convergence (epoch 200), the model demonstrates sharp tumor delineation and high cross-modal consistency, particularly in the FLAIR and T2 modalities, where pathological features are most prominent.

\subsubsection{Comparative Qualitative Analysis}
\autoref{fig:qualitative_grid} illustrates a side-by-side comparison of the synthesized MRI volumes. While the VAE-GAN produces structurally coherent images, they often lack the fine-grained texture and sharp pathology-to-tissue boundaries found in the Ground Truth. The standard GAN baselines (CGAN and 3M-CGAN) exhibit noticeable artifacts and a lack of sharpness in the tumor regions.
In contrast, our proposed ALDM ($K=16, s=3.0$) generates volumes that most closely resemble the ground truth in terms of both 2D slice detail and 3D structural coherence. Specifically, the tumor regions (highlighted in red) in our model maintain consistent intensity profiles across T1, T2, and FLAIR modalities, whereas other models often struggle with \textit{mode collapse} in the high-intensity FLAIR regions. Furthermore, the ablation variants of ALDM ($s=0.3$ to $s=1.0$) exhibit high structural similarity to the default model, confirming that although $s=3.0$ yields the sharpest results, the framework remains robust across various configurations.

%% file: sec/6_conc.tex
\section{Conclusion}
We proposed the ALDM framework, establishing a benchmark for high-fidelity 3D MRI synthesis in data-scarce, few-shot regimes. By transferring anatomical priors from a data-rich GBM domain into a compact latent space, our method achieves superior distributional similarity and structural consistency while remaining robust to hyperparameter variations. The model's clinical utility is validated by high-performing downstream classifiers, proving it captures essential diagnostic features rather than superficial textures. While demonstrating significant potential for augmenting rare disease data, future work will focus on integrating generative-predictive modules and enhancing pathology-aware control.

%% file: sec/7_apdx.tex
\appendix

\section{User Interface}
\begin{figure}[htbp]
    \centering
    \shadowbox{\includegraphics[width=\linewidth]{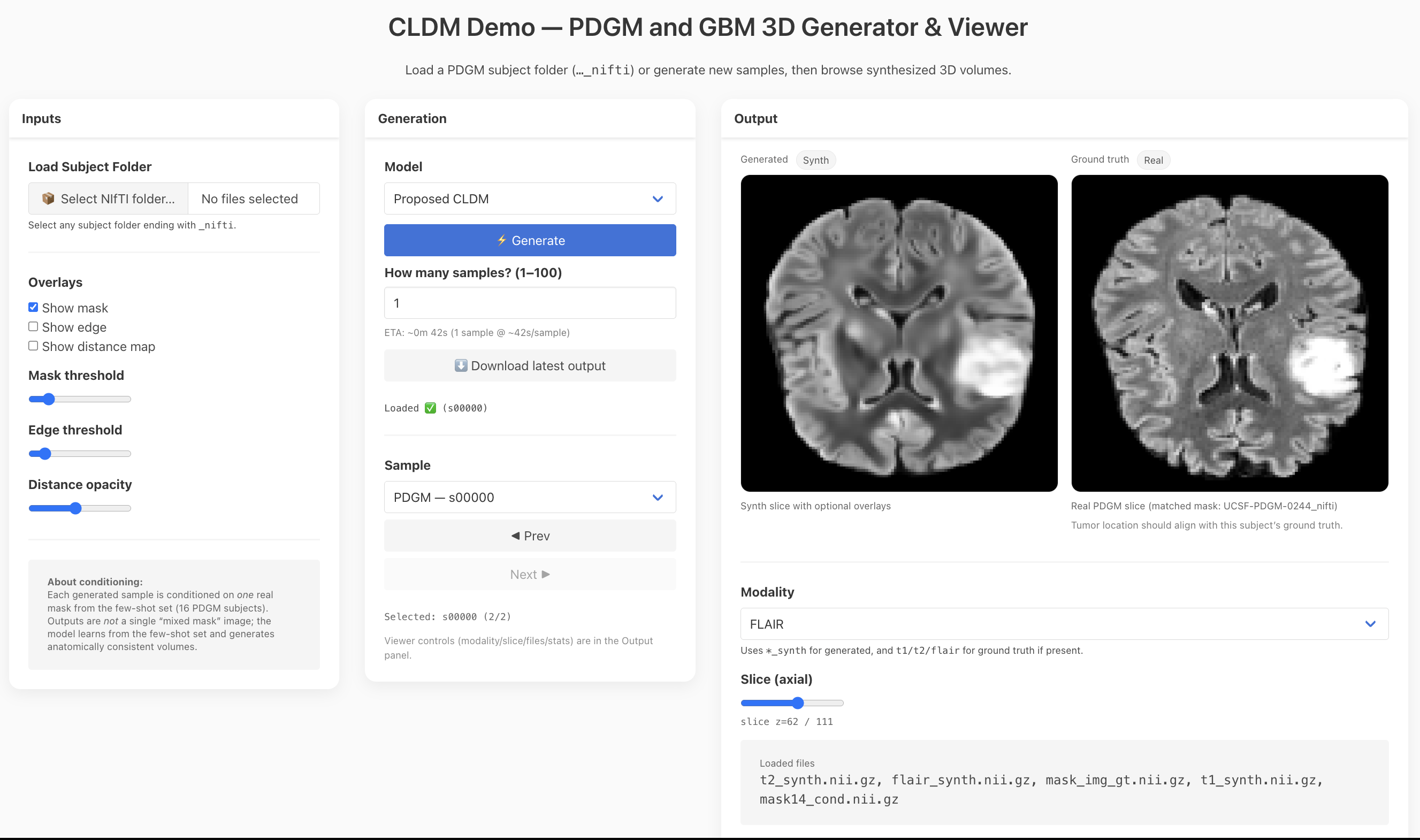}}
    \caption{Web-based user interface for interactive MRI generation and qualitative comparison between ground-truth and synthetic volumes.}
    \label{fig:demo}
\end{figure}

To facilitate qualitative evaluation and interactive exploration of the proposed framework, we developed a web-based user interface that enables on-demand generation of synthetic MRI volumes for both the GBM and PDGM domains\footnote{A demonstration of the user interface is available at \url{https://youtu.be/OARkHCNygmY}.}. As illustrated in \autoref{fig:demo}, the interface allows users to directly compare generated samples against their corresponding ground-truth images, providing intuitive insight into anatomical fidelity, tumor morphology, and cross-domain behavior.

Given the emphasis on few-shot image generation, the interface is designed to highlight both controlled and unconstrained synthesis. Specifically, the first $16$ generated samples reuse tumor masks aligned with the ground-truth images,
resulting in synthetic volumes with tumors located at identical spatial positions. This setting isolates the model’s ability to preserve anatomical structure and intensity characteristics under fixed spatial constraints. All subsequent generated samples reuse the same set of masks but apply them to different anatomical contexts, producing tumors at novel spatial locations and
enabling assessment of the model’s capacity for spatial generalization beyond the observed target-domain examples.

